\documentclass{article}

\PassOptionsToPackage{numbers}{natbib}
\usepackage[preprint]{nips_2018}
\usepackage{macro}

\usepackage{graphicx}
\usepackage{subfigure}
\usepackage{amsmath}
\usepackage{macro}

\usepackage[utf8]{inputenc} 
\usepackage[T1]{fontenc}    
\usepackage{hyperref}       
\usepackage{url}            
\usepackage{booktabs}       
\usepackage{amsfonts}       
\usepackage{nicefrac}       
\usepackage{microtype}      
\title{Cross Domain Image Generation through Latent Space Exploration with Adversarial Loss}

\author{
  Yingjing Lu\\
  Carnegie Mellon University\\
  \texttt{yingjinl@andrew.cmu.edu} \\
}

\begin{document}
\maketitle
\begin{abstract}
  Conditional domain generation is a good way to interactively control sample generation process of deep generative models. However, once a conditional generative model has been created, it is often expensive to allow it to adapt to new conditional controls, especially the network structure is relatively deep. We propose a conditioned latent domain transfer framework across latent spaces of unconditional variational autoencoders(VAE). With this framework, we can allow unconditionally trained VAEs to generate images in its domain with conditionals provided by a latent representation of another domain. This framework does not assume commonalities between two domains. We demonstrate effectiveness and robustness of our model under widely used image datasets. 
\end{abstract}
\section{Introduction}
Humans are can easily learn to transfer knowledge of one domain to another. They can flexibly learn to connect knowledge they already learned in different domains together so that under conditionals within one domain they can recall or activate knowledge they learned from another. Deep generative models are well know for encoding implicit knowledge within one domain through mapping them to latent space. They can be controlled to generate specific samples within learned domain through conditionals. However, compared to humans, deep generative models are less flexible to make new connections from one domain to another. In another word, once it has learned to generate samples  from one set of domain conditionals, making it to adopt to generate samples conditioned on another set of control is often hard and may require to retrain the model, which is often expensive. 

There are many works proposed to address this issue by proposing different approaches to allow deep generative models to transfer knowledge from one to another more flexible. In particular, Engel et al\cite{engel2017latent} recently propose a solution to map conditional encoding to an unconditionally trained VAE to allow it to generate samples conditionally with user defined domain and has achieved excellent results. One limitation is that those conditionals are specifically defined through an one-hot vector. Doing this requires feature engineering and is less effective when we want to condition on some features that are implicit such as using image from one scene as conditional to generate related images in learned domain.   

Another solution is provided by Domain transfer network\cite{taigman2016unsupervised} that trains an end to end model with the assumption of the two domains are somewhat related. In this way, the embedding produced by the autoencoder that encodes images from one domain can be used along with the generated sample to identify whether the transfer learned is effective. To improve on that we tend to develop a framework that make less assumptions between the two domains.

\begin{figure}
  \centering
  \includegraphics[width=0.8\columnwidth]{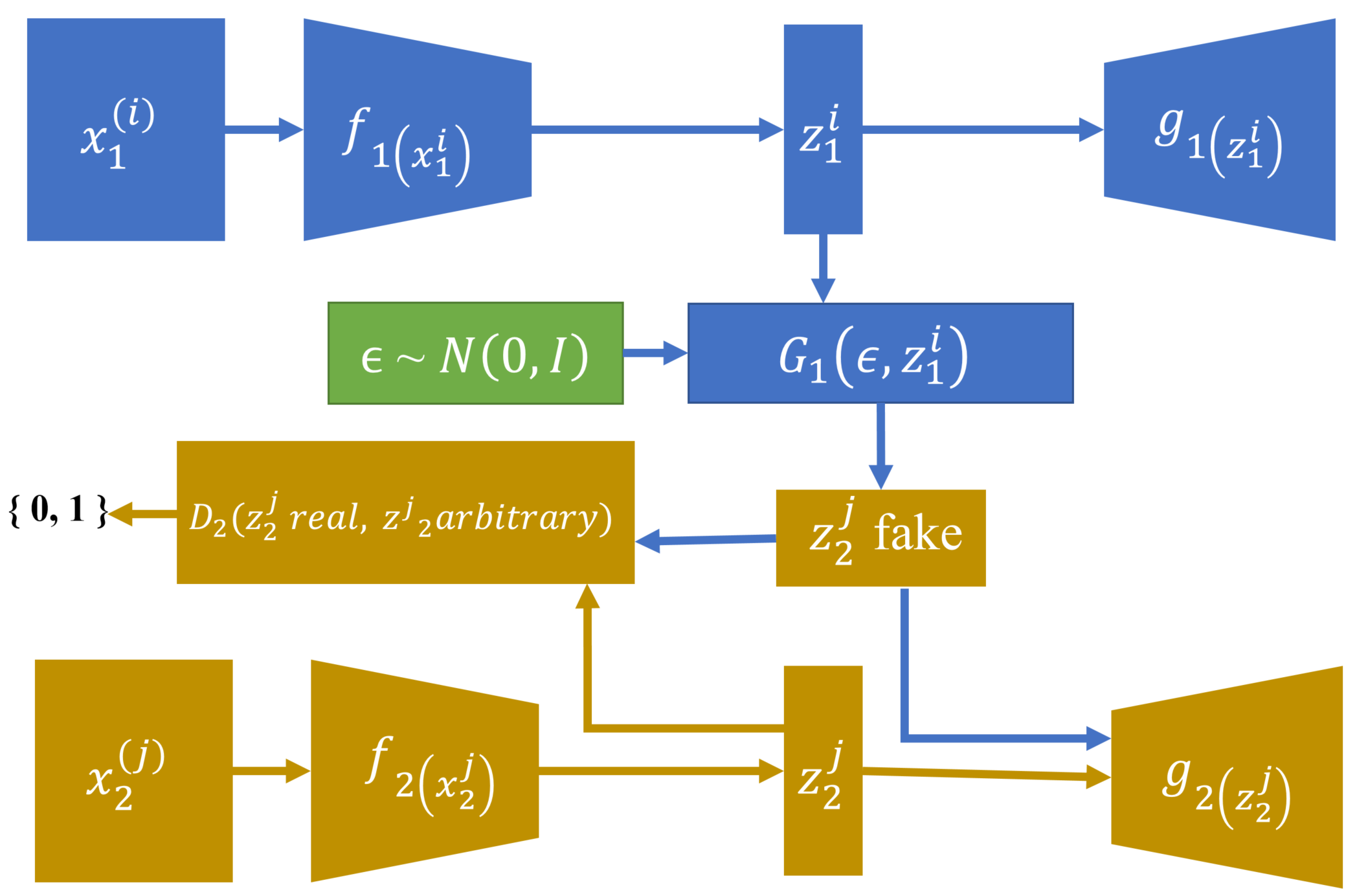}
  \caption{The training process of the generator that transforms latent $z_1^{(i)}$ in class i from domain 1 to $z_2^{(j)}$ in class j from domain 2. For clarity, we only illustrate learning from one direction that is from domain 1 to domain 2. $G_1$ is the generator that takes in latent embedding from domain 1 $z_1^{(i)}$ as conditional and $\epsilon$ sampled from an arbitrary prior to produce an sample $z_2^j ~ fake$ that maps to a possible probability space that can generate samples from class $j$ in domain 2 . We in this case use simple Gaussian noise as prior. The discriminator for domain 2 $D_2(z_2^{(j)}~real, z_2^{(j)}~arbitrary)$ takes in real embedding from class $j$ $z_2^{(j)}$ as conditional to distinguish whether the embedding is generated by true $f_2$ or from $G_1$. Note that the framework is bi-directional. To transform from domain 2 to domain 1 we just need a reverse set of generator and discriminator}
\end{figure}

\section{Methodology}
We will use foot index such as $\mathcal{D}_1$ to define variables from one corresponding domain and head notes such as $x^{(i)}_1$ to define $i$th $x$ within domain 1. We also use bold alphabets to express random variables such as $\xv$. Our proposed framework lies on conditionally generating images from two different domains $\mathcal{D}_1$ with probability distribution $p_1(\xv_1)$, $\mathcal{D}_2$ with probability distribution $p_2(\xv_2)$ for images from domain 1 and domain 2 respectively. $\xv_1$ and $\xv_2$ here being i.i.d within each domain. We define $\gamma_1^{i} \rightarrow \gamma_2^{j}$ to be the user defined  arbitrary condition to generate images in class $j$ from domain 2 under conditionals of images of class $i$ from domain 1. One example would be to generate image of '1' from learned domain 2 conditioned on a image of '2' in domain 1. We make no assumption of any commonalities between the two domains, nor do we assume that the conditional between the two domains involves explicit matching features (such as generating '1' from domain 2 conditioned on an image of '1' from domain 1). In order to map image generation process from two domain with a newly defined domain condition, our framework explores the potential transformation law to map defined class of image from one domain to the corresponding class of image in another domain through latent space. 

Without loss of generality, we define two VAEs $V_1$ and $V_2$ with encoding function $z_1 = f_1(x_1)$, $z_2 = f_2(x_2)$ as encoder functions for each VAE and $g_1(z_1)$ and $g_2(z_2)$ as the corresponding decoder network that are trained unconditionally on the two image domains respectively. We use the word unconditionally to indicate that the two VAEs are both trained with standard unsupervised fashion without any additional conditionals involved in training.The objective function of the two VAEs are formulated by maximizing the estimated lower bound(ELBO)\cite{kingma2013auto}:
\begin{equation}
\log p(x) 
	\geq KL[q_{\phiv}(\zv \mid \xv) \| p_{\thetav}(\zv \mid \xv)] - E_{z\sim q_{\phiv}(\zv \mid \xv)}[\log q_{\phiv}(\zv \mid \xv) - \log p_{\theta}(\xv, \zv)] = ELBO
\end{equation}
Here $p(x)$ represent true distribution of the domain dataset. and encoder and decoder functions are parametrized by $\phi$ and $\theta$ respectively.Thus the total loss function we used in training our VAEs is:
\begin{equation}
\mathcal{L}(\theta, \phi, \xv^{(i)}) = \frac{1}{N} \sum_{i=1}^{N} \lambda_1 \mathcal{C}(\xv^{(i)}, g(f(\xv^{(i)}))) + \lambda_2 ELBO^{(i)}
\end{equation}
Here $\mathcal{C}(\xv^{(i)}, g(f(\xv^{(i)})))$ denotes the pixel-wise reconstruction cost and the $ELBO$ is described above. we here abused the notation of $\{\lambda_1, \lambda_2\}$ to indicate the two hyperparameters that are used to balance the reconstruction cost and the ELBO. The two hyperparameters are shared among the two VAEs throughout our experiments.

With the given cost functions for training we further assume that the two VAEs are well trained in that they can unconditionally reconstruct images with high fidelity given the images within their respective domains. Our training  and implementation details are provided in the experiment section.

To learn a given domain conditional $\gamma_1^{i} \rightarrow \gamma_2^{j}$ we will sample images $x_1^{(i)}$ for arbitrary sample of class $i$ in $\mathcal{D}_1$ and $x_2^{j}$ for arbitrary sample of class $j$ from $\mathcal{D}_2$. We adopt the notion of generative adversarial network(GANs) and use a generator $G(\epsilon, z^{(i)})$ to transform embedding learning from one VAE to the corresponding embedding of another. The generator takes in a noise sampled from a simple prior distribution and the embedding $z^{(i)}$ serves as the conditional. We will follow our presentation structure and use the setting of transforming from domain of $VAE_1$ to domain of $VAE_2$ form class $i$ to class $j$. Thus the generator on domain 1 side generates encoding $z_2^{j}~fake = G_1(\epsilon, z_1^{(i)})$. During training, the result from $G_1$ is passed to the discriminator on the domain 2 side $D_2(z_2^{(j)}, z_2~arbitrary)$. The discriminator from domain 2 takes true encoding from $f_2(x_2^{j})$ as conditional and to determine whether $z_2~arbitrary$ is in the true embedded subspace of class $j$ in domain 2. To make the discriminators stronger within this model we expand our loss function from the 'traditional' GAN loss. For clarity, we introduce $z^{'}$ for the arbitrary embedding z input to the discriminator, and $z$ for the true embedding generated by the encoder network. We shorthand notation $\mathcal{L}_{c=1}(z,z) \equiv -\log(D(z,z))$ for the loss of the true vector under true class, $\mathcal{L}_{c=1}(z,z^{'}) \equiv -(1-\log(D(z, z^{'})))$ as the discriminator loss for classifying a false embedding under true conditional, and finally an additional term $\mathcal{L}_{c=0}(z, \epsilon) \equiv -(1-\log(D(z, \epsilon))$ as the false classification for entering a random noise from a simple distribution. The third term here aims to strengthen the discriminator that under the conditional, the embedding generated from $G$ should not be a random noise vector that seems to comply with the pattern of the distribution. The full loss function of the discriminator is then:
\begin{equation}
    \mathcal{L}_D = \Eb_{z\sim q(z|x)}\big[ \mathcal{L}_{c=1}(z,z) \big] + \Eb_{z\sim G(z,\epsilon)}\big[ \mathcal{L}_{c=0}(z,z^{'}) \big] + \Eb_{\epsilon \sim p(\epsilon)}\big[ \mathcal{L}_{c=0}(z,\epsilon) \big]
\end{equation}

Opposite to discriminator which also restrict from random noise, we introduce a regularization term that is inspired by the log regularization term proposed by \cite{engel2017latent} as $\frac{1}{n}|| \epsilon - G(\epsilon, z)||^2_2$ here $z$ represents the input embedding to the generator to be transformed. Intuitively, as the generator shift the simple noise to the mapped distribution, it would usually maximize the distance between the embedding generated and the original noise. With this term added in minimizing the loss function will create a force that "pulls back" $G$ from moving from $\epsilon$ too far, thus encourage variety of the embedding generated. Adding the regularization term resulting in out generator loss:
\begin{equation}
    \mathcal{L}_G = \Eb_{z\sim p(z), \epsilon \sim p(\epsilon)}\big[ \mathcal{L}_{c=1}G(z,\epsilon) + \frac{\lambda_{reg}}{n}|| \epsilon - G(\epsilon, z)||^2_2 \big]
\end{equation}

During training, we first train two VAEs until convergence, then we train two pairs of generator and discriminator alternatively until convergence. During sampling, we sample from desired class $i$ in domain 1 and feed in $VAE_1$ then VAE will encode the sample to embedding $z_1^{(i)}$ and transformed by generator with a noise $G_1(\epsilon, z_1^{(i)})$, this new transformed embedding will be passe to decoder of $VAE_2$ to generate actual sample $g_2(G_1(\epsilon, z_1^{(i)}))$.

\section{Experiments}
We performed our experiments mainly on MNIST\cite{lecun-mnisthandwrittendigit-2010} and SVHN\cite{netzer2011reading} datasets. MNIST dataset contains 28X28 hand written digits from 0 to 9 in grey scale with approximately 60000 training samples and 20000 testing samples. SVHN dataset contains digit photos captured in street cropped to 32X32 in RGB. Training set contains more than 73000 images and testing set contains more than 26000 images for digits from 1 to 9.

For \textbf{MNIST to MNIST} mapping, we split the training set and testing set with number as their classes. To adhere to our assumption and avoid feature matching, we assign domain 1 contain digits $\{0,1,2,3,4\}$ and domain 2 contains digits $\{5,6,7,8,9\}$. The conditional generation law is defined to generate a certain class of digit given a class of digit. For conditional generation from domain 1 to domain 2 we define $\{0 \rightarrow 5, 1 \rightarrow 6, 2 \rightarrow 7, 3 \rightarrow 8, 4 \rightarrow 9\}$ and conditional generation from domain 2 to domain 2 is the opposite direction. 

To evaluate the accuracy quantitatively, we train a convolutional neural network on the full train set of MNIST that serve as a classifier for evaluating whether our image generated belongs to the right class. Our convolutional neural network achieve 99.2\% classification accuracy on MNIST test set so that we are confident that our classifier would be a good fit to evaluate whether our model can generate sample in the right class. Aside from the classifier we also consider the recognizablility of a digit. If a digit is generated too vague to be identified or mixed between digits, we as humans regard those samples as false positives as they are not valid generation and wrongly projected in latent space. We ask three different volunteers to eyeball a subset of the image generated and record their classification result as a comparison. To avoid unknown effect of one digit can perform extremely well to conditionally generate another digit, we shuffle the conditional pair 3 times and report our results based on the average performance recorded from the three conditional pairs. 

For \textbf{SVHN to MNIST} mapping we use similar scheme to regard digits as different classes. Since SVHN does not contain digit 0, we drop the pair 0-5 and only uses $\{ 1 \rightarrow 6, 2 \rightarrow 7, 3 \rightarrow 8, 4 \rightarrow 9\}$ as our major experiment target and shuffle pairs among those. Since we cannot train a classifier that can perform classification of generated SVHN images with high accuracy(the best classification accuracy we had was 66.8\%), we use MNIST digits generated conditioned on SVHN images to report the performance. Similarly, we use both trained convolutional neural network and human as classifier to evaluate the results quantitatively. 

For all of our experiments, we train our models on training set and report results from images generated from testing set. 
\begin{figure}
  \centering
  \includegraphics[width=\columnwidth]{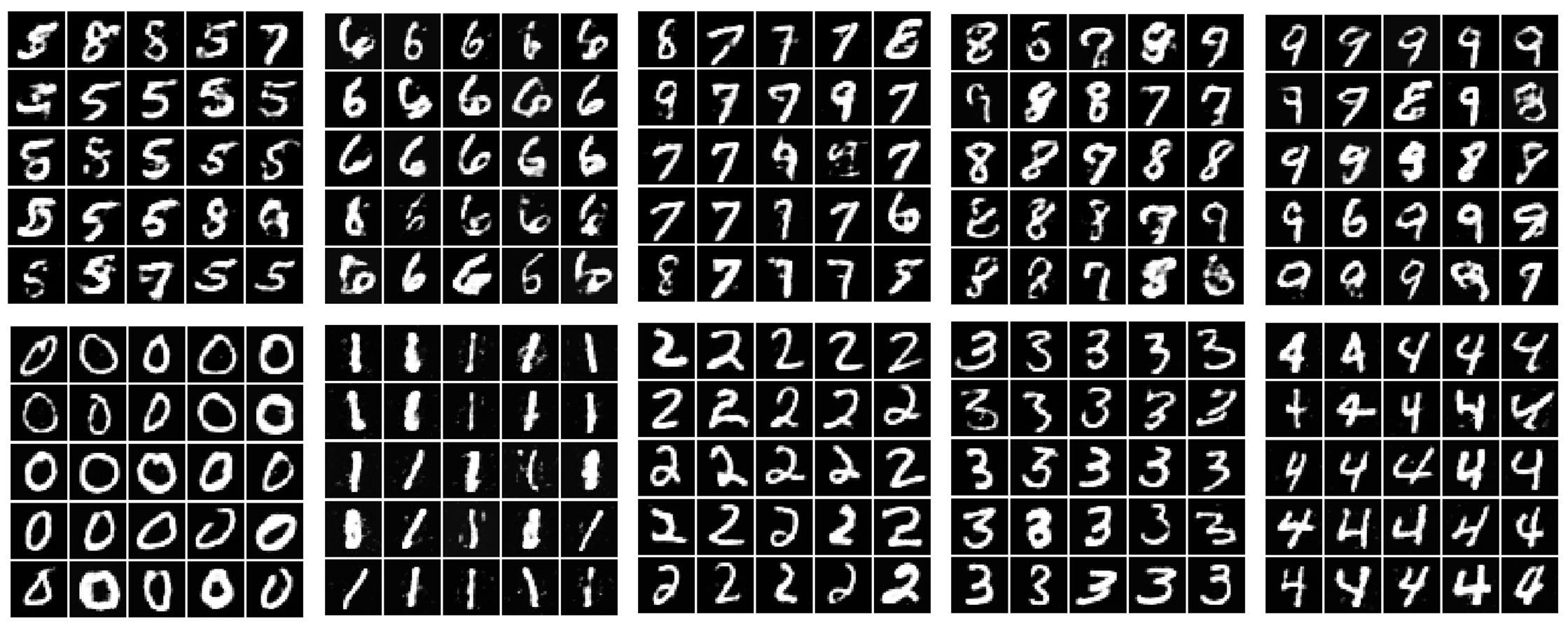}
  \caption{MNIST to MNIST generation with the conditional pairs described above. The model is trained with full train set and tested on the random part of the test set. Bottom is the image generated from the reconstruction, i.e. the conditionals. Top are the corresponding class of images generated conditioned on the bottom domain. We observe that most of the results are generated accurately and with high fidelity.The variant of style is obvious so it is confident to say that the generator did not experience severe model collapse.}
\end{figure}
\begin{figure}
  \centering
  \includegraphics[width=\columnwidth]{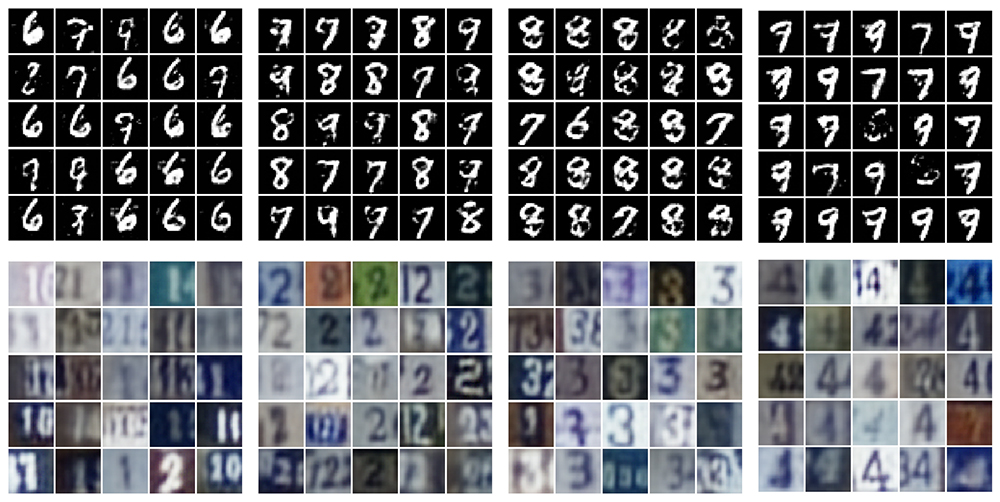}
  \caption{SVHN to MNIST generation with the conditional pairs described above. The model is trained with full train set and tested on the random part of the test set. Bottom is the image generated from the reconstruction, Top are the corresponding class of images generated conditioned on the bottom domain. The SVHN is a lot more complicated. Multiple numbers appearing in the image and color channel disrupt the distribution and result in a more mixed generator distribution.}
\end{figure}
\begin{figure}
  \centering
  \includegraphics[width=0.8\columnwidth]{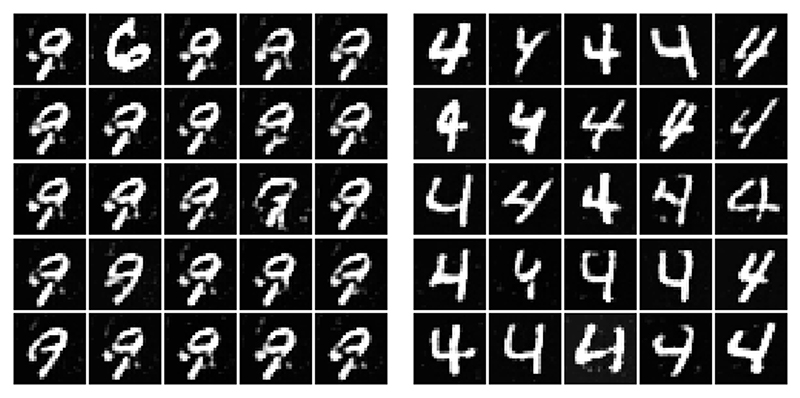}
  \caption{MNIST to MNIST generation. We observe that if the $L_2$ regularization term is removed from the generator loss function, training becomes less stable and resulting in mode collapse more often.}
\end{figure}
\subsection{Quantitative results on image generation}
From the quantitative perspective, we see that our framework generally perform well on MNIST dataset transformation. Probably due to MNIST's simple distribution. The average accuracy of the CNN classifier reports higher than the average accuracy reported by human volunteers. We inspected and found that the classifier tend to be lenient when the sample is vague and tend not to classify to the right class where as human volunteers tend not to guess which class it belongs and simply report it wrong. Especially with small training size the samples have a lot of jitters and volunteers tend not to guess what the digit actually is. SVHN to MNIST yields less accuracy. From Figure 3 we see that the SVHN contains many disruptive samples that make the conditional less inclined to a particular class. Combining the results from the two datasets together we see that degradation of accuracy resulting in decreased training set is relatively small, suggesting that our model is relatively robust in the datasets we examined.
\begin{table}
  
  \caption{Accuracy of generated images}
  \label{sample-table}
  \centering
  \begin{tabular}{llll}
    \toprule
    \cmidrule(r){1-4}
    Dataset     & Size of training set     & Accuracy(CNN classifier)   &Accuracy(Human) \\
    \midrule
    MNIST $\rightarrow$ MNIST & 500       & 0.63    &  0.58   \\
    MNIST $\rightarrow$ MNIST & 1000       & 0.68    &   0.65  \\
    MNIST $\rightarrow$ MNIST & 2000       & 0.66    &  0.67   \\
    MNIST $\rightarrow$ MNIST & Full set       & 0.77    &  0.75   \\
    \midrule
    SVHN $\rightarrow$ MNIST  & 500       & 0.43    &    0.48\\
    SVHN $\rightarrow$ MNIST  & 1000       & 0.48    &  0.44\\
    SVHN $\rightarrow$ MNIST  & 2000       & 0.46    &  0.43\\
    SVHN $\rightarrow$ MNIST  & Full set       & 0.61    &  0.58 \\
    \bottomrule
  \end{tabular}
\end{table}
\subsection{Implementation details}
For the we uses symmetric architecture for the two variational autoencoders. we use three layer convolutional neural networks followed by a linear layer of size 256. Z embedding is set to 100 for all experiments. For the decoder we use 1 linear layer of size 2048 followed by  4 layers of deconvolution layers. We apply RELU on all layers except the output uses tanh and batch normalization on convolutional and deconvolutional layers. We use a hyperparameter $\alpha = 0.1$ as the coefficient applied to the $\sigma$ outputed from the encoder. This hyperparameter is used by Engel et.al.\cite{engel2017latent} as the authors in the this paper show that imposing this hyperparameter allow the distribution of the embedded space to be tighter.

For the generator and discriminator pair, we use a paired 4-layer fully connected network of size 512. The conditional and the noise $\epsilon$ was simply concatenated and feed to the first layer. The generator outputs a transformed embedding as well as a gating factor resulting from sigmoid of the transformed embedding. The final output transformed embedding is an interpolation between the input random noise $\epsilon$ and the transformed embedding with gateing factor as the interpolation coefficient. This structure is also introduced by \cite{engel2017latent}. We adopt this architecture and believe that this interpolation can introduce more variety of output.  
\section{Related works}
Generative adverserial networks or GAN for short, is well known for generating realistic samples through mapping latent manifolds with noise and conditionals inputed into generator(s).Many prior works has illustrates successful attempts to disentangle how GAN encode features into latent space and conditional GANs(cGAN) is a particular class of GANs that are proven to control output contents through concatenating tractable conditionals with selected noise. In the conditional setting, a lot of previous work has been done to explore possibilities to control samples generated from GANs through encoded conditionals\cite{mirza2014conditional,reed2016generative,isola2017image}.Here in particular, Zhu et. al.\cite{NIPS2017_6650} proposed a way to transform style of image from one domain to another through a cycle encoding and verification structure and has achieved decent results. We view the BiGAN proposed by Donahue ei.al.\cite{donahue2016adversarial} as a concurrent work of ours in that it also explores the concept of conditional noise mapping. It would be interesting to see if the model can be adaptable to the setting not being trained end-to-end. The notion of approximating an implicit distribution is not restricted to GAN alone and many works have focused on incorporating adversarial loss to achieve more flexible latent representation approximation\cite{makhzani2015adversarial,dumoulin2016adversarially,mescheder2017adversarial}. 

Autoencoder(AE) is another class of deep generative model that compiles inputs and reconstruct samples through encoding features in latent space. Recent work done by Liu et.al.\cite{liu2017unsupervised} has shown that VAE can perform well on matching the latent space between images from two domains. The added adversarial loss facilitates the training of the VAE by preventing it from mode collapsing. 

\begin{figure}
  \centering
  \includegraphics[width=\columnwidth]{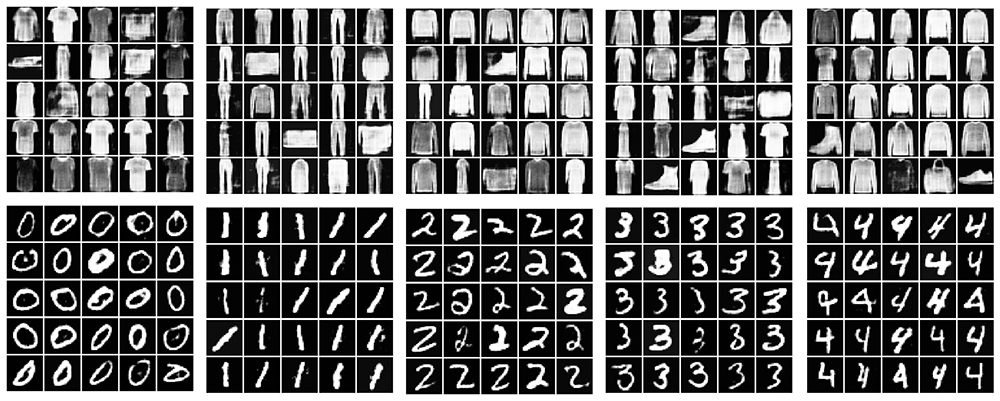}
  \caption{MNIST to FASHION-MNIST\cite{xiao2017/online} generation. Labeling correlation is 0 - T-shirt, 1-Trouser, 2-Pullover, 3-Dress, 4-Coat. Bottom is the original MNIST reconstruction, top is the corresponding image generated using bottom as conditional mapping.}
\end{figure}

\begin{figure}
  \centering
  \includegraphics[width=\columnwidth]{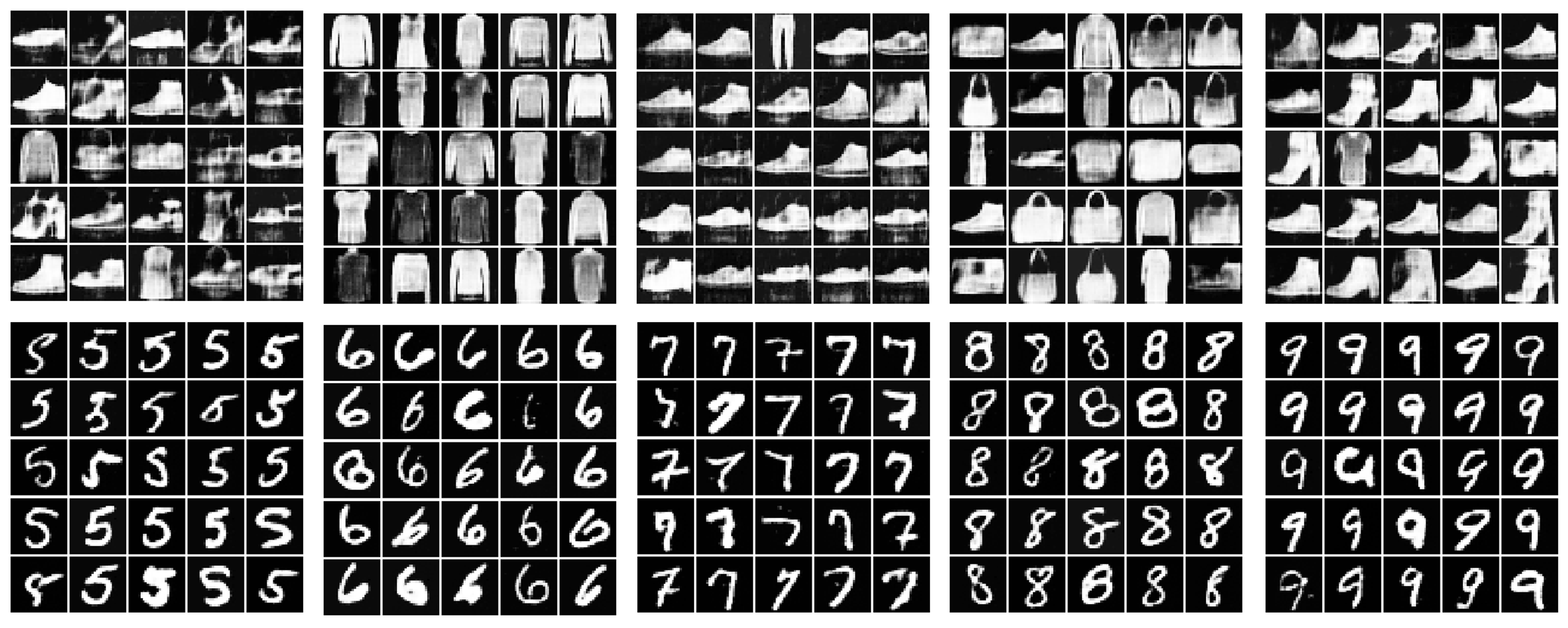}
  \caption{MNIST to FASHION-MNIST\cite{xiao2017/online} generation. Labeling correlation is 5-Sandal, 6-	Shirt, 7-Sneaker, 8-Bag, 9-Ankleboot. Bottom is the original MNIST reconstruction, top is the corresponding image generated using bottom as conditional mapping.}
\end{figure}

\section{Discussion and future work}
In this work we present a framework that can decouple the latent space of the variational encoder and match the arbitrary domains through adversarial training. To shift the conditional generation from one domain to another, the only thing to do is to train a separate pair of generator and discriminator that can transfer latent features from one domain to another. No need to retrain the entire VAE. Both qualitative and quantitative results have shown that this approach produces promising results and is easy to implement. In the future we will explore this framework on more complex datasets such as CIFAR-10\cite{krizhevsky2009learning} or Oxford Flower\cite{Nilsback08}. Those datasets either contain smaller images or with more complex distributions. Under those conditions a variational autoencoder may not be the best option to provide a clear domain boundary for discriminators in our framework to distinguish. But our framework is not limited to variational autoencoders and can be applied to other models with latent space embeddings. We will explore more generative models that can incorporate our framework in the future. 

\small
\bibliography{biba}
\bibliographystyle{plainnat}

\end{document}